\documentclass[letterpaper]{article} 
\usepackage[]{aaai24}  
\usepackage{times}  
\usepackage{helvet}  
\usepackage{courier}  
\usepackage[hyphens]{url}  
\usepackage{graphicx} 
\usepackage{natbib}  
\usepackage{caption} 
\usepackage{algorithm}
\usepackage{algorithmic}
\usepackage{amsmath}
\usepackage{amssymb}
\usepackage{float}
\usepackage{forest}
\usepackage{booktabs}
\usetikzlibrary{shadows}
\usepackage[inline]{enumitem}
\usepackage{array}    

\usepackage{color}

\definecolor{lightgray}{gray}{0.9}

\usepackage{newfloat}
\usepackage{listings}
\DeclareCaptionStyle{ruled}{labelfont=normalfont,labelsep=colon,strut=off} 
\floatstyle{ruled}
\newfloat{listing}{tb}{lst}{}
\floatname{listing}{Listing}
\usepackage{xcolor} 
\usepackage[most]{tcolorbox}  
\usepackage{mdframed}
\mdfdefinestyle{KeyTakeaway}{
    linecolor=gray,
    outerlinewidth=2pt,
    roundcorner=10pt,
    innertopmargin=7pt,
    innerbottommargin=7pt,
    innerrightmargin=7pt,
    innerleftmargin=7pt,
}

\mdfdefinestyle{Box1}{style=KeyTakeaway, backgroundcolor=blue!10}
\mdfdefinestyle{Box2}{style=KeyTakeaway, backgroundcolor=green!10}
\mdfdefinestyle{Box3}{style=KeyTakeaway, backgroundcolor=red!10}
\mdfdefinestyle{Box4}{style=KeyTakeaway, backgroundcolor=orange!10}
\mdfdefinestyle{Box5}{style=KeyTakeaway, backgroundcolor=purple!10}
\mdfdefinestyle{Box6}{style=KeyTakeaway, backgroundcolor=pink!10}
\mdfdefinestyle{Box7}{style=KeyTakeaway, backgroundcolor=lime!10}
\mdfdefinestyle{Box8}{style=KeyTakeaway, backgroundcolor=cyan!10}
\newcommand{\primarykeywords}{
(1) Large Language Models;
(2) Learning;
(3) Planning
}

\title{The Case for Developing a Foundation Model for Planning-like Tasks from Scratch}

 \author {
     Biplav Srivastava, Vishal Pallagani
 }
\affiliations{
    Artificial Intelligence Institute, University of South Carolina\\
     \{biplav.s@, vishalp@mailbox.\}sc.edu
}

\begin{document}

\maketitle
    
\begin{abstract}
Foundation Models (FMs) have revolutionized many areas of computing including Automated Planning and Scheduling (APS).  For example,  a recent study found them to be useful for eight aspects of planning problems: plan generation, language translation, model construction, multi-agent planning, interactive planning, heuristics optimization, tool integration, and brain-
inspired planning. Besides APS, there are a number of seemingly related tasks involving generation of a series of actions with varying guarantee of their executability to achieve intended goals, that we collectively call {\em planning-like} (PL) tasks like business processes, programs, workflows, and guidelines, where researchers have considered using FMs. However, previous works have primarily focused on  pre-trained, off-the-shelf FMs and optionally fine-tuned them. In this paper, we discuss the need for a comprehensive FM for PL tasks from scratch and explore the design considerations for it. We argue that such a FM will open new and efficient avenues for PL problem solving just like LLMs are creating for APS.

\end{abstract}


\section{Introduction}
\label{sec:intro}

\begin{figure}
 \centering
  \includegraphics[width=0.45\textwidth]{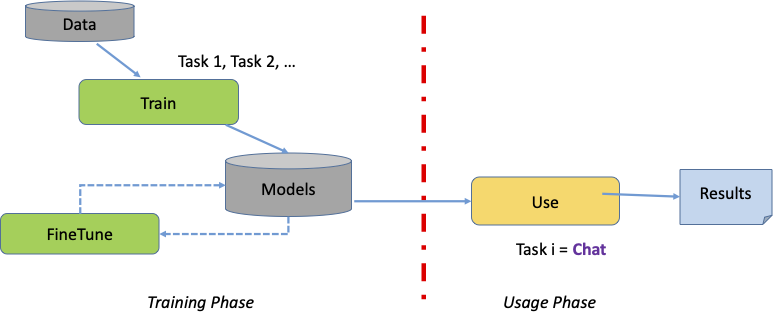}
  \caption{Overview of developing an FM-based system - e.g., an LLM-based chatbot.}
  \label{fig:chatbot-llm}
\end{figure}

Foundation Models (FMs \cite{github-fm-survey}), or Large Language Models (LLMs \cite{github-llm-survey}) when referring to them in textual format, have revolutionized many areas of computing.
We will use the terms FM and LLM   inter-changeably to refer to models like GPT  \cite{openai2023gpt4}, PaLM \cite{chowdhery2022palm}, LLaMA \cite{touvron2023llama} and ImageBind \cite{girdhar2023imagebind}.
These 
are deep-learning based models trained using large datasets on a set of tasks (see Figure~\ref{fig:chatbot-llm}), and optionally fine-tuned using datasets of a particular domain. An FM model so trained can be used for a variety of tasks including conversation (chat), i.e., dialog management. ChatGPT and Bard are examples of such chatbots.  When testing such a model, neither the training procedure, consisting of data or tasks, may be known nor the reason why the chatbot generated an utterance - 
thus, they are truly {\em blackbox} systems.


\begin{figure}
 \centering
  \includegraphics[width=0.45\textwidth]{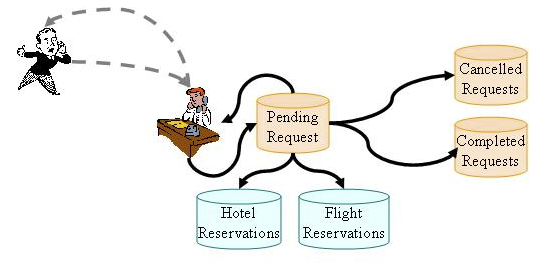}
  \caption{The travel agency example from \cite{srivastava2003web}}
  \label{fig:travel-agency}
\end{figure}

Automated Planning and Scheduling (APS) is a branch of Artificial Intelligence (AI)  that focuses on the creation of action sequences, also called plans or policies,  for execution by intelligent agents. 
 Besides APS, a number of other tasks involve generating a series of actions with varying guarantee for execution semantics, that we collectively call {\em planning-like} (PL) tasks including business processes \cite{omg2011bpmn}, dialogs \cite{dialogdataset-2015serbansurvey}, guidelines \cite{field1990clinical-guidelines}, instructions, design diagrams \cite{cad_Wu_2021_ICCV}, programs \cite{kernighan88-clang,cacm-copilot-study}, workflows \cite{workflow-book}. For example, a recent instruction generation task for travel planning \cite{xie2024travelplanner-nlp} was introduce in the Natural Language Processing (NLP) community that can be processed by humans. But the  application scenario of booking
travel packages in a travel agency was standardized as back as in 2002 \footnote{See
http://www.w3.org/2002/04/17-ws-usecase.}  (Figure~\ref{fig:travel-agency}) and spans from a simple,
closed-world travel example into a dynamic, integrated
solution. 


\begin{figure}
 \centering
  \includegraphics[width=0.45\textwidth]{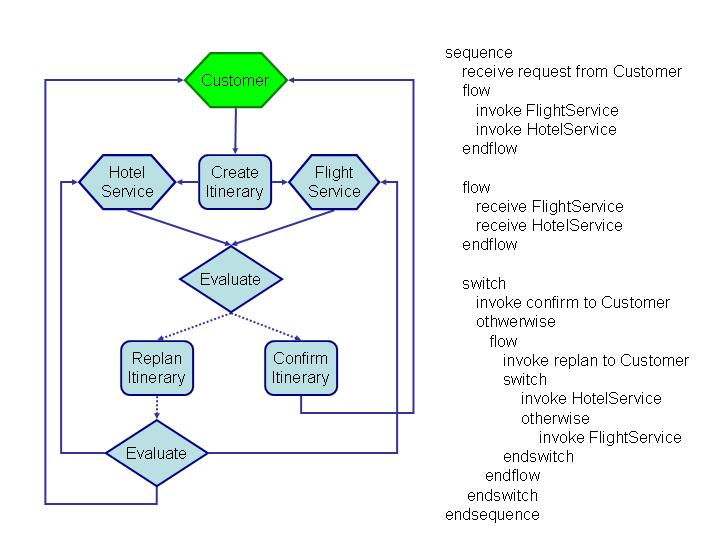}
  \caption{The travel plan in business process notations, from \cite{srivastava2003web}}
  \label{fig:travel-agency-bpel}
\end{figure}

In the {\em closed-world} case, a customer talks to the travel agent
who notes the customer's requests and generates a {\em trip request}
document that may contain several needed {\em flight} and {\em hotel
reservations}. The travel agent performs all bookings and when he is
done, he puts the {\em trip request} either into the {\em cancelled
requests} or the {\em completed requests} data base. A completed document is sent to
the customer as an answer to his request. If the booking fails, the
customer is contacted again and the whole process re-iterates. 
Now if we allow  the travel agency  to cooperate with external
specialized service providers that offer hotel and flight
reservations, the process requires to reorganize the entire processing of customer
requests. In the {\em open-world} case now, new services have to be integrated and all services must
correctly interact with each other. 
 Different settings of travel planning were popularly tackled in APS \cite{agarwal2008understanding}, business processes literature (see Figure~\ref{fig:travel-agency-bpel}), and web services \cite{srivastava2003web}. 


 Researchers have started considering FMs for PL tasks extensively.
 In APS, a recent study found LLMs to be relevant for various aspects of planning problems: plan generation, language translation, model construction, multi-agent planning,
interactive planning, heuristics optimization, tool integration,
and brain-inspired planning \cite{icaps2024-llmplanning-prospectssurvey}.  Their use has also lead to many ongoing debates. Consider plan generation.
Although there is growing consensus that LLMs cannot reason but act as approximate information retrievers \cite{valmeekam2023planning,valmeekam2023planbench,valmeekam2022large,valmeekam2023can}, they can still be useful to find plans under various LLM/ fine tuning settings \cite{pallagani2022plansformer,pallagani2023plansformer,pallagani2023understanding}. Furthermore, they  can be  productively used in conjunction with conventional planners for effective plan generation \cite{fabiano2023fast}.


However, most of the previous works have used pre-trained, off-the-shelf FMs and optionally fine-tuned them. 
In this paper, we discuss the need for a comprehensive FM for PL tasks from scratch and explore the design considerations for it. We argue that such a FM will open new and efficient avenues for PL problem solving just like LLMs are creating for APS.    
We begin by providing background on FMs, APS and planning-like tasks. Then, we outline the desiderata for a Planning FM and its benefits over current models. We then discuss design considerations, data choices and training options, and conclude.

\section{Background}

In this section, we give relevant background for classical planning problem in APS, PL tasks, FMs and LLMs,  and LLM training so that the motivation and design options for a ground-up FM for PL is well contextualized.

\subsection{Automated Planning and Scheduling}

The simplest APS task is a classical planning problem (CPP).
A CPP is a tuple \( \mathcal{M} = \langle \mathcal{D}, \mathcal{I}, \mathcal{G} \rangle\) with domain \( \mathcal{D} = \langle F, A \rangle \) - where \(F\) is a set of fluents that define a state \( s \subseteq F\), and \(A\) is a set of actions - and initial and goal states \( \mathcal{I}, \mathcal{G} \subseteq F\) \cite{aimodern}. Action \( a \in A\) is a tuple \( (c_a, \textit{pre}(a), \textit{eff}^{\pm}(a)) \) where \( c_a \) is the cost, and \( \textit{pre}(a), \textit{eff}^{\pm}(a) \subseteq F\) are the preconditions and add/delete effects, i.e., \( \delta_{\mathcal{M}}(s, a) \models \bot s \) \textit{if} \( s \not\models \textit{pre}(a); \) \textit{else} \( \delta_{\mathcal{M}}(s, a) \models s \cup \text{eff}^{+}(a) \setminus \text{eff}^{-}(a) \) where \( \delta_{\mathcal{M}}(\cdot) \) is the transition function. The cumulative transition function is \( \delta_{\mathcal{M}}(s, (a_1, a_2, \ldots, a_n)) = \delta_{\mathcal{M}}(\delta_{\mathcal{M}}(s, a_1), (a_2, \ldots, a_n)) \). A plan for a CPP is a sequence of actions \( \langle a_1, a_2, \ldots, a_n \rangle \) that transforms the initial state \( \mathcal{I} \) into the goal state \( \mathcal{G} \) using the transition function \( \delta_{\mathcal{M}} \). Traditionally, a CPP is encoded using a symbolic representation, where states, actions, and transitions are explicitly enumerated. This symbolic approach, often implemented using Planning Domain Definition Language or PDDL \cite{mcdermott1998pddl}, ensures precise and unambiguous descriptions of planning problems. This formalism allows for applying search algorithms and heuristic methods to find a sequence of actions that lead to the goal state, which is the essence of the plan.


\subsection{ Planning Like Tasks}
 

We now clarify terminology for prominent tasks that bear resemblance to planning and we seek to include them in the proposed FM. They are summarized in Table~\ref{tab:characterizing-pl}.

The most general representation is a {\bf workflow} which is a formalized representation of activities  involved in accomplishing a well-defined objective using automated and manual actions consisting of control and/or data flow \cite{workflow-book}. 
So, workflows can represent manufacturing operations ({\bf design}), machine learning pipelines, complex operations in a financial institution, {\bf dialogs} generated by chatbots, or scripts written for automation. 
In the context of {\em automated} action execution, they are referred to as {\bf plans} or {\bf policies} in AI community or {\bf programs} in the software engineering community, 
and in the context of business improvement, they are also called {\bf business processes}. When tasks are described for managing or communicating with people, they are also represented as {\bf instructions} or {\bf guidelines}.




\begin{table*}[t!]
\centering
\scriptsize
\caption{Characterizing Select Planning Like Tasks. Auto Gen: Automatic Generation, Auto Exec: Automatic Execution}
\label{tab:characterizing-pl}
\resizebox{\textwidth}{!}{%
\begin{tabular}{@{}p{1.5cm}p{0.5cm}p{0.8cm}p{0.8cm}p{0.8cm}p{0.8cm}p{3.5cm}p{1.8cm}@{}}
\toprule
\textbf{Name} & \textbf{State} & \textbf{Control Flow} & \textbf{Data Flow} & \textbf{Auto Gen} & \textbf{Auto Exec} & \textbf{Comments} & \textbf{Reference}\\
\midrule

Business Process 
&            
& x                      
& x                  
&                   
&                    
& Data, failure, compensation conditions and cross-organization responsibilities are richly represented; execution can be manual          
& \cite{omg2011bpmn}\\

Design drawing       
&                
& x                      
&                   
&                   
&                   
& Multi-modal data, geometry constraints    
& \cite{cad_Wu_2021_ICCV}\\ 

Dialogs       
&                
& x                      
&                   
&  *                 
&                   
& Multi-agent conversation with at least one human, *: trust issues are prominent    
& \cite{dialogdataset-2015serbansurvey}\\ 

Guidelines       
&                
& x                      
&                   
&                   
&                   
& Focused on normative (domain) preferences and constraints     
& \cite{field1990clinical-guidelines}\\ 

Instructions       
&                
& x                      
&                   
&                   
&                      
& Step-by-step sequence of actions, emphasizing simplicity and clarity for interpretation and action.
& \cite{weitzenhoffer1974instruction}\\ 

Plan 
& x
& x
& x
& x
& x
& Full initial state, partial goal state, minimal data representation, execution guarantee (soundness)
& \cite{mcdermott1998pddl} \\

Program       
&                
& x                      
& x                  
&                   
& x                  
& States, including goals, are not formally represented.    
& \cite{kernighan88-clang}\\ 

Web Services
&                
& x                      
& x                  
& *                  
& x                  
& Web-based endpoints, *: with semantic descriptions added        
& \cite{wsdl2,agarwal2008understanding} \\

Workflow
&                
& x                      
& x                  
&                   
& x                  
& Data, failure conditions and handling, compensations elaborately represented          
& \cite{workflow-book} \\

\bottomrule
\end{tabular}
}
\end{table*}


Thus, in summary:
\begin{itemize}

\item Business Processes: a set of coordinated activities described within a business context across organizations - e.g., customers and vendors \cite{omg2011bpmn,srivastava2010apqc,srivastava2010business}.

\item Design drawings: Computer-aided design (CAD) diagrams represent a sequence of operations for manufacturing new physical items. They can have both text and visual representations and now FMs are being considered for them \cite{cad_Wu_2021_ICCV,vuruma2024cloud}.

\item Dialogs: represent interaction between a human and automated system with the collective goal of solving a problem \cite{dialogdataset-2015serbansurvey,dialog-intro}, and are a popular task with both FMs \cite{jalil2023chatgpt} and application of planning \cite{dialog-planning-adi,dialog-planning-muise,Pallagani2021AGD-prudent}. Trust issues are a key concern with them \cite{dialog-ethics,apollo-chatbots}.

\item Guidelines: a sequence of expert-defined information consisting of questions and constraints (e.g. clinical practice guidelines) \cite{field1990clinical-guidelines}.

\item Instructions: a sequence of activities a human can follow to achieve a desirable goal \cite{weitzenhoffer1974instruction}.

\item Plans: a set of coordinated activities where states and action models are formally represented, guaranteeing soundness of execution for any plan found.  
\item Programs: a set of functions implementing activities and executable on a computer (computable platform) \cite{kernighan88-clang}; Copilot is a popular FM-based tool that has been shown to improve coding productivity \cite{cacm-copilot-study}.

\item Web Services: a set of functions run on computers connected over a network (e.g., internet) \cite{srivastava2003web,wsdl2}.

\item Workflows: a structured sequence of activities and tasks involving automated systems and human participants designed to achieve a specific goal or process efficiently.


\end{itemize}

We observe from Table~\ref{tab:characterizing-pl}  that control flow is the common information across PL tasks. However, since other information may be missing, the workflows - referring to the PL outputs collectively - may be acceptable for humans but not for automation.
Ensuring automated systems precisely interpret and execute complex, multi-step tasks requires understanding task-specific actions and contextualizing these within the physical world ({\bf grounding}). It also demands alignment with objectives and adaptability to various contexts ({\bf alignment}), ensuring coherent and efficient progression towards goals. Additionally, the ability to dynamically adapt to new instructions in changing environments ({\bf instructability}) is crucial, necessitating advanced capabilities to bridge the gap between formal task representations and real-world execution. While intrinsic to APS, developing these capabilities in FMs is an active research area. 




\subsection{Foundation and Large Language Models}

Large language models are neural network models with upwards of $\sim$ 3 billion parameters trained on extremely large corpora of natural language data (trillions of tokens/words). These models are proficient in interpreting, generating, and contextualizing human language, leading to applications ranging from text generation to reasoning tasks. The Transformer \cite{vaswani2017attention} architecture, which is fundamental to advancements in language modeling, has undergone modifications to develop LLMs that are adept at a diverse range of tasks. Three notable architectural variants in language modeling include causal, masked, and sequence-to-sequence models.

\noindent \textbf{Causal Language Modeling (CLMs)}: CLMs, such as GPT-4 \cite{openai2023gpt4}, are designed for tasks where text generation is sequential and dependent on the preceding context. They predict each subsequent word based on the preceding words, modeling the probability of a word sequence in a forward direction. This process is mathematically formulated as shown in Equation \ref{eqn:clm}.

\begin{equation}
    P(T) = \prod_{i=1}^{n} P(t_i | t_{<i})
    \label{eqn:clm}
\end{equation}

In this formulation, $P(t_i | t_{<i})$ represents the probability of the $i$-th token given all preceding tokens, $t_{<i}$. This characteristic makes CLMs particularly suitable for applications like content generation, where the flow and coherence of the text in the forward direction are crucial.

\noindent \textbf{Masked Language Modeling (MLMs)}: Unlike CLMs, MLMs like BERT\cite{vaswani2017attention} are trained to understand the bidirectional context by predicting words randomly masked in a sentence. This approach allows the model to learn both forward and backward dependencies in language structure. The MLM prediction process can be represented as Equation \ref{eqn:mlm}.

\begin{equation}
    P(T_{\text{masked}}|T_{\text{context}}) = \prod_{i \in M} P(t_i | T_{\text{context}})
    \label{eqn:mlm}
\end{equation}

Here, $T_{\text{masked}}$ is the set of masked tokens in the sentence, $T_{\text{context}}$ represents the unmasked part of the sentence, and $M$ is the set of masked positions. MLMs have proven effective in NLP tasks such as sentiment analysis or question answering.

\noindent \textbf{Sequence-to-Sequence (Seq2Seq) Modeling}: Seq2Seq models, like T5 \cite{2020t5}, are designed to transform an input sequence into a related output sequence. They are often employed in tasks that require a mapping between different types of sequences, such as language translation or summarization. The Seq2Seq process is formulated as Equation \ref{eqn:seq2seq}.

\begin{equation}
    P(T_{\text{output}}|T_{\text{input}}) = \prod_{i=1}^{m} P(t_{\text{output}_i} | T_{\text{input}}, t_{\text{output}_{<i}})
    \label{eqn:seq2seq}
\end{equation}

In Equation \ref{eqn:seq2seq}, $T_{\text{input}}$ is the input sequence, $T_{\text{output}}$ is the output sequence, and $P(t_{\text{output}_i} | T_{\text{input}}, t_{\text{output}_{<i}})$ calculates the probability of generating each token in the output sequence, considering both the input sequence and the preceding tokens in the output sequence.

In addition to their architectural variants, the utility of LLMs is further enhanced by model adaptation strategies. One key strategy is fine-tuning, which involves further training pre-trained LLMs on a smaller, task-specific dataset, thereby adjusting the neural network weights for particular applications. This process is mathematically represented in Equation \ref{eqn:fine-tuning}.

\begin{equation}
    \theta_{\text{fine-tuned}} = \theta_{\text{pre-trained}} - \eta \cdot \nabla_{\theta}L(\theta, D_{\text{task}})
    \label{eqn:fine-tuning}
\end{equation}

Here, $\theta_{\text{fine-tuned}}$ are the model parameters after fine-tuning, $\theta_{\text{pre-trained}}$ are the parameters obtained from pre-training, $\eta$ is the learning rate, and $\nabla_{\theta}L(\theta, D_{\text{task}})$ denotes the gradient of the loss function $L$ with respect to the parameters $\theta$ on the task-specific dataset $D_{\text{task}}$.

In contrast to fine-tuning, in-context learning offers a distinct adaptation strategy for LLMs, notably seen in CLMs. This method allows models to tailor responses to immediate prompts without additional training. Given a context $C$, the model generates text $T$ that is contextually relevant, as shown in Equation \ref{eqn:icl}. Here, $P(T|C)$ is the probability of generating text $T$ given the context $C$, and $P(t_i | t_{<i}, C)$ is the probability of generating the $i$-th token $t_i$ given the preceding tokens $t_{<i}$ and the context $C$. 


\begin{equation}
    P(T|C) = \prod_{i=1}^{n} P(t_i | t_{<i}, C)
    \label{eqn:icl}
\end{equation}

\begin{figure*}[htbp]
    \centering
    \includegraphics[scale=0.31]{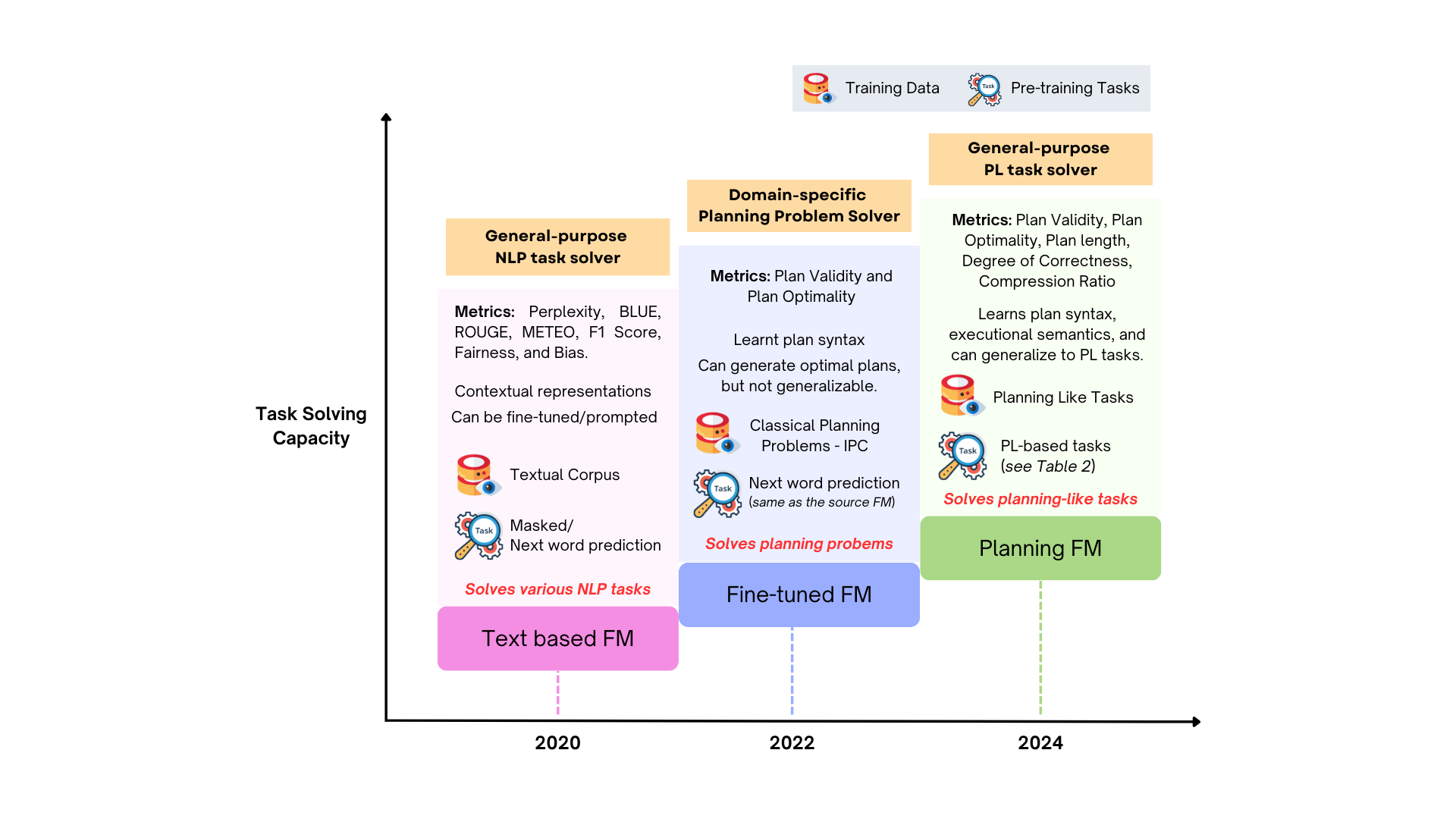}
    \vspace{-0.9cm}
    \caption{A timeline capturing the progression of FMs for planning and the need for designing a Planning FM from ground up for solving PL tasks. }
    \label{fig:timeline}
\end{figure*}

\noindent {\bf Which model to start with for plan generation?}
\citep{pallagani2023understanding} identify that pre-trained LLMs cannot generate plans for classical planning problems off the shelf. The selection of pre-training data and the methodology employed for model adaptation play pivotal roles in improving the quality of plan generation. It is observed that LLMs pre-trained on programming languages, as opposed to solely natural language, when fine-tuned with incrementally hard planning problems \cite{plansformer-paper-pallagani2022}, exhibit enhanced capabilities in generating optimal plans for domains in which they have been fine-tuned. Despite these advancements, the fine-tuned models demonstrate a limitation in their ability to generalize to domains not included in the fine-tuning process.

\subsection{Training LLMs}

There are tutorials on training CLMs, Masked, and Seq2Seq LLMs for NLP from scratch, including \cite{ak-gpt-fromscratch} and \cite{train-llm-scratch-hf}. Training LLMs requires a vast and high-quality dataset, as the data's quality directly influences model performance. With adequate data, LLMs can achieve near-human proficiency across various domains \cite{gunasekar2023textbooks}. Training data typically involves diverse sources such as Books, CommonCrawl, Reddit feeds, Wikipedia, and Code. While next-word prediction remains the primary pre-training task for LLMs, there is a shift towards incorporating domain-specific pre-training tasks. Such tasks, designed for specialized fields like code synthesis and Chess strategy analysis, offer more targeted learning pathways for the models.

For planning, there is a recent tutorial on fine-tuning LLMs
\cite{tutorial-llmplanning-2024harnessing} with planning problems for plan generation. Additionally, there has been progress in adapting the Transformer architecture to A\textsuperscript{*} search dynamics for puzzles like Sokoban \cite{lehnert2024beyond}. However, a significant gap persists in the availability of a comprehensive training corpus for PL tasks and novel pre-training tasks apt for an FM to learn the execution semantics and the syntactic knowledge intrinsic to a PL task.

\section{The Need for a Planning Foundation Model}
\label{sec:desiderata}
\begin{table*}[htbp]
\centering
\small
\caption{Novel Pre-training Tasks for PL Tasks}
\label{tab:pretraining-tasks}
\begin{tabular}{@{}>{\raggedright\arraybackslash}p{0.18\textwidth} >{\raggedright\arraybackslash}p{0.51\textwidth} >{\raggedright\arraybackslash}p{0.20\textwidth}@{}}
\toprule
\textbf{Pre-training Task} & \textbf{Description} & \textbf{Relevance} \\
\midrule
Next Action Prediction & Training the model to make context-based decisions, simulating step-by-step plan execution. & Mimics real-world planning and workflow execution. \\
\addlinespace
Conditional Branching Prediction & Predicting outcomes from multiple possible branches in a given scenario. & Reflects decision-making in complex processes. \\
\addlinespace
Action and Effect Modeling & Understanding causal relationships between actions and their consequences. & Fundamental for realistic plan generation. \\
\addlinespace
Constraint Satisfaction & Identifying and applying constraints to optimize outcomes. & Crucial for generating efficient and viable plans. \\
\addlinespace
Hierarchical Task Planning & Generating plans involving tasks at multiple levels of abstraction. & Addresses complex tasks with sub-tasks in detailed planning. \\
\addlinespace
Cross-Domain Understanding & Recognizing entities and relationships across different planning domains. & Enhances model's versatility, applicability, and generalizability. \\
\addlinespace
Execution Simulation & Simulating the execution of plans or code to predict outcomes. & Improves the model's execution semantics understanding. \\
\addlinespace
Error Detection & Identifying and correcting errors in plans or instructions. & Enhances reliability and correctness of outputs. \\
\addlinespace
Multi-modal Contrastive Learning & Incorporating visual, textual, and auditory data for plan understanding and generation. & Relevant for tasks relying on multi-modal inputs. \\
\bottomrule
\end{tabular}
\end{table*}
Current approaches to FMs in NLP and AI research have predominantly centered around generic pre-training tasks such as masked or next-word/sentence prediction, as shown in Figure \ref{fig:timeline}. These models are trained on diverse text corpora, enabling them to perform various downstream tasks, from sentiment detection to question answering. However, when these models are applied to PL tasks, their effectiveness is inherently limited by their generic training paradigms. This limitation is due to the complex requirements of PL tasks, which not only necessitate the understanding of textual information but also require an intricate capture of state, control flow, and data flow alongside the constrained generation and execution semantics unique to each task (see Table \ref{tab:characterizing-pl}). Current pre-training tasks lack the specificity to model these detailed, dynamic relationships inherent in PL tasks, leading to a gap in effectiveness when transitioning from general text understanding to specialized plan generation and execution.


The finetuning of FMs, exemplified by models like Plansformer, represents a step towards domain-specific adaptation, where generic FMs are trained on PL tasks. While this method offers improvements in performing PL tasks by leveraging domain-specific data, it still falls short in fully capturing the nuances of plan generation, validation, and generalizing across all PL tasks due to the limitation in training data and the inherent general-purpose pre-training. Prompting is another approach that has gained popularity recently. Despite its wide adoption, it has demonstrated inferior plan generation performance compared to fine-tuning and often relies heavily on the human-in-the-loop's skill in crafting effective prompts.

The need for a Planning FM arises from these limitations and new opportunities. Note from Table~\ref{tab:characterizing-pl} that PL tasks have a complementary focus, which may be leveraged with an FM. For example, business processes and dialogs focus on multi-party interaction, design focuses on geometry; plans focus on soundness, and workflows and business processes focus on failure handling. An FM trained on diverse considerations may bring these insights learned from data to bear in its output. It can also focus on PL-specific tasks during both pre-training and downstream and metrics to enable better execution of semantics associated with PL tasks. Such an FM may be critical for effectively generating, summarizing, and generalizing PL tasks. We note that in the past, in code generation, FMs trained for that objective \cite{wang2021codet5} performed better than fine-tuning text-based FMs \cite{raffel2020exploring} with code - a trend seen in many other areas.

\section{Discussion}
\label{sec:discussion}

We now discuss how the proposed FM may be developed.
The envisioned Planning FM distinguishes itself through a specialized focus on PL tasks and its design principles of compactness, generalizability, and an intrinsic awareness of temporal and execution semantics.

\subsection{Training Procedure}
We provide a detailed description of the comprehensive training procedure involved in constructing the Planning FM, encompassing critical aspects from tokenization and model architecture to the application of the Planning FM in various downstream tasks.

\noindent \textbf{Tokenization and Embeddings:} 
Designing a tokenizer for the Planning FM, which is specialized for handling PL tasks requires a nuanced approach to understand and encode diverse data types and structures. Such a tokenizer can be designed as follows:

\begin{itemize}
    \item \textbf{Pre-processing and Normalization}
    \begin{itemize}
        \item Input Normalization: The initial stage involves standardizing heterogeneous inputs, including textual data from various PL task documents and non-textual data such as design drawings. Textual data undergo normalization processes, including case normalization, special character removal, and spelling correction. Non-textual data is processed through Optical Character Recognition (OCR) techniques to extract textual information, ensuring a unified text-based input for further processing.
        \item Semantic Augmentation: Employing domain-specific ontologies such as the Planning Ontology, the text is augmented with semantic annotations. This involves the identification and annotation of domain-relevant entities and their interrelations, enhancing the model's comprehension of planning-specific lexicon and structures.
    \end{itemize}

    \item \textbf{Tokenization Strategy}
    \begin{itemize}
        \item Adaptive Subword Tokenization: A subword tokenization algorithm, tailored to the planning domain through the training on the pre-processed and normalized corpus, is employed. Techniques such as Byte-Pair Encoding (BPE), Unigram Language Model, or SentencePiece can be considered, with the aim to capture the lexical characteristics of PL tasks.
        \item Multi-Modal Tokenization: For incorporating non-textual information, a multi-model tokenization approach is adopted. This involves the extraction of visual features using Convolutional Neural Networks (CNNs), subsequently mapped into a discrete token space compatible with textual tokens, facilitating integrated multi-modal analysis.
    \end{itemize}

    \item \textbf{Special Tokens and Embeddings}
    \begin{itemize}
        \item Integration of Domain-Specific Tokens: The tokenizer incorporates special tokens designed to signify critical PL constructs (e.g., START\_OF\_PLAN, END\_OF\_TASK) and entities (e.g., OBJECTS, COST). These tokens are crucial for the model to recognize and prioritize PL task elements within the data.
        \item Enhanced Embedding Layer: The embedding layer for the Planning FM incorporates rotary position embeddings (RoPE)\cite{su2024roformer} to effectively encode sequential information in PL tasks. RoPE distinguishes itself by directly encoding positional information into the embeddings, allowing the model to maintain the relative order of tokens without losing the context of their placement.
    \end{itemize}
\end{itemize}

This approach to tokenizer for Planning FM provides for a blend of linguistic processing, domain-specific tailoring, and multi-modal integration, enabling the model to be adept at PL tasks.

\noindent \textbf{Model Architecture}
A Seq2Seq architecture will be the preferred choice  for the Planning FM, as prior research has demonstrated its capabilities in generating structured sequences compared to CLMs or MLMs. The Seq2Seq framework, originally devised for machine translation, excels at comprehending and producing complex sequences, making it an optimal choice for PL tasks.

The Seq2Seq architecture comprises two principal components: an encoder and a decoder. The encoder processes the input sequence, capturing its semantic and structural essence into a comprehensive context vector. For PL tasks, this involves encoding the initial state, objectives, constraints, and available resources. The inclusion of a self-attention mechanism enables the encoder to assess the relative importance of different elements within the input sequence, a critical function for understanding complex planning instructions and dependencies.

The decoder is responsible for generating the output sequences, such as plans or summaries, from the encoded representation. It predicts the subsequent action or step in the plan sequentially, considering both the current generation and the overarching goal articulated by the encoder. The decoder employs a cross-attention mechanism, particularly allowing it to concentrate on pertinent portions of the input sequence while formulating each step of the plan. A significant advancement in the Seq2Seq architecture for the Planning FM will be the incorporation of RoPE. RoPE enhances the Planning FM's ability to maintain and exploit the temporal and sequential dependencies more effectively than the traditional positional encoding methods utilized in modern LLMs.

\noindent \textbf{Pre-training Objectives}
The novel pre-training objectives for the Planning FM are outlined in Table \ref{tab:pretraining-tasks}. We depart from conventional next-word prediction objective predominantly used to train LLMs to objectives that are inherently aligned with the intricacies of PL tasks. Our proposed pre-training objectives encompass tasks such as Next Action Prediction, Conditional Branching Prediction, and Action-Effect Modeling, equipping the Planning FM to learn context-based decision-making, understanding complex process dynamics, and modeling causal relationships between actions and their consequences. Furthermore, tasks like Constraint Satisfaction and Hierarchical Task Planning target the generation of optimal and executable plans by navigating constraints and orchestrating tasks across multiple levels of abstraction. The inclusion of Cross-Domain Understanding and Execution Simulation extends the model's versatility and applicability across various PL scenarios, enhancing its predictive accuracy and execution semantics. Additionally, Error Detection and Multi-modal Contrastive Learning further refine the model's reliability and adaptability to multi-modal inputs, ensuring the generation of robust and error-free plans. 
\noindent \textbf{Evaluation}
The training regimen of the Planning FM, featuring novel pre-training tasks specifically devised for PL tasks, necessitates the introduction of sophisticated evaluation metrics to accurately gauge its efficacy. These metrics aim to capture essential dimensions of the model's output, such as \emph{Plan Validity}, which scrutinizes the feasibility and compliance of the generated plans with predefined constraints and objectives. \emph{Plan Optimality} assesses the efficacy and resource efficiency of the plans, ensuring goals are met with minimal expenditure of time and resources. \emph{Plan Length}, defined as a numerical value representing the number of actions or steps in a generated plan. \emph{Degree of Correctness} measures the ratio of successfully achieved goals to the number of specified goals within a plan. Notably, \emph{Compression Ratio} is introduced as a metric to assess the model’s efficiency in condensing detailed planning instructions into succinct, actionable summaries without losing critical information. This suite of metrics offer a comprehensive framework for assessing the Planning FM's performance, ensuring it delivers practical, optimal, and executable plans suited to a wide array of real-world planning scenarios.

\noindent \textbf{Downstream Tasks}
The Planning FM is designed for a variety of downstream tasks, reflecting its utility in planning-related applications. Key downstream tasks include:
\begin{itemize}
    \item Plan Generation: Creating detailed plans based on specific goals and constraints.
    \item Completing a Partial Plan: Filling in missing elements of an incomplete plan to ensure its comprehensiveness and actionability.
    \item Replanning: Adjusting plans in response to changes in objectives, constraints, or conditions.
    \item Predicting Plan Validity: Determining the feasibility and coherence of plans, identifying potential execution issues.
    \item Plan Summarization: Condensing comprehensive plans into brief summaries that retain essential information \cite{process-workflow-summary}.
    \item Resource Optimization: Ensuring efficient use of resources within plans to achieve objectives with minimal waste.
    \item Error Detection and Correction: Identifying and correcting inaccuracies or inconsistencies within plans.
\end{itemize}

The model can be further adapted to any downstream PL task by fine-tuning or prompting the Planning FM with with an appropriate quantity and quality of data.
\subsection{Model Properties}

We envisage the proposed FM to be compact, general and aware of PL needs. We now describe specific steps towards the same.

\noindent \textbf{Compactness:} Conventional FMs are assumed to be usable in a resource-rich setting like residing remotely on a cloud and invoked on-demand via API calls. 
However, many real world applications are resource-constrained and we there is a rising demand for compact models that can be deployed effectively in both cloud  as well as edge settings \cite{vuruma2024cloud}  without compromising performance.

Prominent techniques for these are:
\begin{itemize}
    \item
    \textbf{Model Pruning:} The process of removing non-critical and redundant components of a model without a significant loss in performance. With respect to LLMs, this can mean removing weights with smaller gradients or magnitudes and parameter reduction among others. Novel pruning methods like Wanda \cite{sun2023simple} and LLM-Pruner \cite{ma2023llmpruner} present optimal solutions for making LLMs smaller. Furthermore, parameter sharing across different parts of the model and pruning redundant or non-contributory neurons post-training contribute to a compact yet powerful model architecture. 

    \item
    \textbf{Quantization:} A key strategy is mixed precision training, which utilizes a combination of 16-bit (half-precision) and 32-bit (single precision) floating-point operations during the training process or representing model parameters such as weights in a lower precision, i.e. using fewer bits to store the value \cite{gholami2021survey}. This results in a smaller model size, faster inference, and a reduced memory footprint. LLM Quantization can be achieved either in the post-training phase  \cite{dettmers2022llmint8} or during the pre-training or fine-tuning phase \cite{liu2023llmqat}.
    
    \item
    \textbf{Knowledge Distillation:} Transferring the knowledge of a large teacher model to a smaller learner model to replicate the original model's output distribution difference. Knowledge Distillation has been widely used to reduce LLMs like BERT into smaller distilled versions DistilBERT \cite{sanh2020distilbert}. More recently, approaches like MiniLLM \cite{gu2023knowledge} and \cite{hsieh2023distilling} further optimize the distillation process to improve the student model's performance and inference speed.
\end{itemize}

\noindent \textbf{Generalizability}: refers to the ability to generate output beyond the specific training data. 
We seek to attain it across the wide range of PL tasks 
by training the Planning FM on a diverse corpus that encompasses not just traditional planning datasets like the International Planning Competition (IPC \cite{ipc-plan-competitions}) but also extends to other tasks like business processes \cite{bpmn-dataset-camunda}, dialogues \cite{dialogdataset-2015serbansurvey}, design \cite{cad_Wu_2021_ICCV},  and workflows \cite{workflow-dataset-plusone} form Table~\ref{tab:characterizing-pl}. Note that the business process and design data are inherently multi-modal consisting of image and text. It  can be further enriched by knowledge from a planning ontology that captures inter-relationships between different metadata \cite{planning-ontology}. This broad training base, combined with task-agnostic pre-training objectives such as next-action prediction, conditional branching prediction, and execution simulation, equips the Planning FM with a robust foundation of knowledge. The ontological knowledge facilitates the model's ability to adapt to various planning contexts without requiring extensive task-specific retraining. Moreover, the incorporation of transfer learning techniques allows the model to leverage knowledge acquired from one task to improve performance on related tasks, enhancing its generalizability. \cite{torrey2010transfer}.

\noindent \textbf{Awareness of temporal and execution considerations} is cultivated by deliberately designing pre-training tasks that mimic real-planning scenarios as outlined in Table \ref{tab:pretraining-tasks}. With the help of novel pre-training tasks, the Planning FM develops a nuanced comprehension of how plans unfold over time and how individual actions interrelate. Training the model to recognize and predict the impact of various actions within different execution contexts further ensures that the generated plans are theoretically sound and practically executable.



\subsection{LLM Properties}

The development of Planning FM introduces notable concerns regarding the executional guarantees traditionally offered by APS systems. FMs inherently do not provide sound and complete solutions due to their probabilistic frameworks. Therefore, leveraging FM architectures for PL tasks requires a thorough investigation into the Planning FM's properties of alignment, instructability, and grounding - areas where FMs often encounter limitations.

\noindent \textbf{Grounding:} refers to the model's ability to base its planning and reasoning processes on real-world knowledge and data. This necessitates the incorporation of extensive domain-specific knowledge into the model, potentially with the help of a knowledge graph \cite{sheth2022process}. The APS planners may also help in verifying physical constraints just as information retrievers do in RAG \cite{rag-grounding-nlp}.  

\noindent \textbf{Alignment:} is critical for ensuring that the progression of Planning FM towards goals is both coherent and efficient. It involves the model's capability to accurately interpret and execute PL tasks in a manner that not only meets predefined objectives but also dynamically adjusts to changes in the environment.

\noindent \textbf{Instructability:} involves the model's responsiveness to nuanced directives and its capacity to adapt its planning strategies accordingly. Enhancing the Planning FM's instructability requires fine-tuning with preferences \cite{ziegler2019fine} that encompass a broad range of PL tasks and user intents.

\section{Conclusion}

In this paper, we considered various PL tasks together and argued for the need for a FM to be developed from scratch. We explored various design
considerations including training tasks, data and its behavioral properties. We argue that such a FM will open new
and efficient avenues for PL problem solving just like LLMs
are creating for APS.

\section{Acknowledgements}

We wish to thank Amitava Das, Kaushik Roy, Manas Gaur, Jianhai Su and Sai Vuruma  for discussions related to foundation models and NLP, and Faez Ahmed on the role of FM for design. This work was partially funded with a gift from JPMorgan Chase Faculty Research Award.


\bibliography{combined}

\end{document}